\documentclass[11pt]{article} 
\usepackage[utf8]{inputenc}
\usepackage{rldmsubmit,palatino}
\usepackage[english]{babel}
\usepackage{amsmath}
\usepackage{amssymb}
\usepackage{mathtools}
\usepackage{amsfonts}
\usepackage{hyperref}
\usepackage[acronym,abbreviations]{glossaries-extra}
\usepackage{glossaries-prefix}
\usepackage[capitalise]{cleveref}
\usepackage{textcomp}
\usepackage{tikz}
\usepackage{float}
\usepackage{array}
\usepackage{import}
\usetikzlibrary{arrows,automata}
\usepackage{algorithm}
\usepackage{algpseudocode}
\usepackage{algorithmicx}
\usepackage{bbm}
\usepackage{xparse}
\usepackage{pgfplots}
\usepackage{pgfplotstable}
\usepackage{wrapfig}
\usepackage{graphicx}
\usepackage{subcaption}
\usepgfplotslibrary{fillbetween}
\usepackage{xcolor}
\usepackage{autonum}

\usepackage{microtype}

\cslet{blx@noerroretextools}\empty
\usepackage[style=numeric,sorting=none,sortcites,citestyle=numeric-comp]{biblatex}

\DeclarePairedDelimiterX{\infdivx}[2]{[}{]}{%
    #1\delimsize\|\;#2%
}
\newcommand{\kl}{D_{\text{KL}}\infdivx}

\newcommand{\psym}{p}
\newcommand{\Psym}{P}
\newcommand{\qsym}{q}
\newcommand{\Qsym}{Q}
\newcommand{\pssym}{\psym^\ast}
\newcommand{\qssym}{\qsym^\ast}

\NewDocumentCommand\pr{mmo}{%
    \ensuremath{%
        #1\mathchoice{\!}{\!}{}{}\left( #2 %
        \IfNoValueTF{#3}{}{\,\middle\lvert\, #3} %
        \right)%
    }
}

\NewDocumentCommand\prs{ommo}{%
    \ensuremath{\pr{\IfNoValueTF{#1}{#2}{#2_{#1}}}{#3}[#4]}
}

\NewDocumentCommand\p{omo}{\prs[#1]{\psym}{#2}[#3]}
\NewDocumentCommand\ps{omo}{\prs[#1]{\pssym}{#2}[#3]}
\NewDocumentCommand\PP{omo}{\prs[#1]{\Psym}{#2}[#3]}
\NewDocumentCommand\q{omo}{\prs[#1]{\qsym}{#2}[#3]}
\NewDocumentCommand\Q{omo}{\prs[#1]{\Qsym}{#2}[#3]}
\NewDocumentCommand\qs{omo}{\prs[#1]{\qssym}{#2}[#3]}

\newcommand{\pc}[2]{\p{#1}[#2]}

\newcommand{\qp}[2]{\q[#2]{#1}}

\newcommand{\uqcp}[3]{\q[#3]{#1}[#2]}

\newcommand{\qcp}[3]{\q[#3]{#1}[#2]}

\newcommand{\misym}{\text{MI}}
\newcommand{\lisym}{\text{LI}}
\NewDocumentCommand\mi{mmo}{\pr{\misym}{#1, #2}[#3]}
\NewDocumentCommand\li{mmo}{\pr{\lisym}{#1, #2}[#3]}

\newcommand{\E}{\mathbb{E}}
\newcommand{\Ex}[2]{\E_{#1}\mathchoice{\!}{\!}{}{}\left[ #2 \right]}

\addto\extrasenglish{%
    \crefname{approx}{Approximation}{Approximations}%
  \Crefname{approx}{Approximation}{Approximations}%
  \creflabelformat{approx}{#2(#1)#3}
}

\addto\extrasenglish{%
  \crefname{constr}{Constraint}{Constraints}%
  \Crefname{constr}{Constraint}{Constraints}%
  \creflabelformat{constr}{#2(#1)#3}
}

\addto\extrasenglish{%
  \crefname{def}{Definition}{Definitions}%
  \Crefname{def}{Definition}{Definitions}%
  \creflabelformat{def}{#2(#1)#3}
}

\newcommand{\vfesym}{\mathcal{F}}

\newcommand{\efesymP}{G}

\pgfplotsset{
    every axis plot/.append style={line width=0.8pt},
    every axis plot post/.append style={
        every mark/.append style={mark=none}
    }
}

\definecolor{pltBlue}{HTML}{1f77b4}
\definecolor{pltOrange}{HTML}{ff7f0e}
\definecolor{pltGreen}{HTML}{2ca02c}
\definecolor{pltRed}{HTML}{d62728}
\definecolor{pltPurple}{HTML}{9467bd}
\definecolor{pltBrown}{HTML}{8c564b}
\definecolor{pltPink}{HTML}{e377c2}
\definecolor{pltGray}{HTML}{7f7f7f}
\definecolor{pltBeige}{HTML}{bcbd22}
\definecolor{pltLightBlue}{HTML}{17becf}

\NewDocumentCommand\plotWithStddev{mmmmmmmO{1}}{\plotstd{#1}{#2}{#3}{#4}{#5}{#6}[#7][#8]}

\NewDocumentCommand\plotstd{mmmmmmoO{1}}{%
    \pgfplotstableread[col sep=comma]{#1}\datatable
    \IfNoValueTF{#7}{}{%
        \addplot[smooth, opacity=0, name path=A, forget plot] table [x=#5, y expr=#3 * (\thisrow{#6} - #8 * \thisrow{#7})] {\datatable};
        \addplot[smooth, opacity=0, name path=B, forget plot] table [x=#5, y expr=#3 * (\thisrow{#6} + #8 * \thisrow{#7})] {\datatable};
        \addplot[#2, opacity=0, fill opacity=0.4]  fill between [of=A and B];
        \addlegendentry{$\pm$ std};
    }
    \addplot+[color=#2, smooth, solid] table [x=#5, y expr=#3 * \thisrow{#6}] {\datatable};
    \addlegendentry{#4};
}

\NewDocumentCommand\plotstdcom{mmmmmmoO{1}}{%
    \pgfplotstableread[col sep=comma]{#1}\datatable
    \IfNoValueTF{#7}{}{%
        \addplot[smooth, opacity=0, name path=A, forget plot] table [x=#5, y expr=#3 * (\thisrow{#6} - #8 * \thisrow{#7})] {\datatable};
        \addplot[smooth, opacity=0, name path=B, forget plot] table [x=#5, y expr=#3 * (\thisrow{#6} + #8 * \thisrow{#7})] {\datatable};
        \addplot[#2, opacity=0, fill opacity=0.4, forget plot]  fill between [of=A and B];
    }
    \addplot+[color=#2, smooth, solid] table [x=#5, y expr=#3 * \thisrow{#6}] {\datatable};
    \addlegendentry{#4};
}

\newcolumntype{C}{>{\centering\arraybackslash}m{0.14\linewidth}}

\pgfplotsset{rewplot/.style={
    align=center,
    legend style={nodes={scale=0.6, transform shape}},
    grid=both,
    width=0.45\textwidth,
    height=5cm,
    reverse legend,
    legend pos=north west,
    legend cell align={left},
    scaled x ticks=false,
    xlabel = Number of episodes
}}

\newcommand{\anaacr}[3]{%
    \newacronym[prefixfirst={a\ },prefix={an\ }]{#1}{#2}{#3}%
}

\anaacr{fep}{FEP}{Free Energy Principle}
\newacronym{vfe}{VFE}{Variational Free Energy}
\newacronym{efe}{EFE}{Expected Free Energy}
\anaacr{fef}{FEF}{Free Energy of the Future}
\newacronym{gfe}{GFE}{Generalized Free Energy}
\anaacr{feef}{FEEF}{Free Energy of the Expected Future}
\newacronym{ai}{AI}{Active Inference}
\newacronym{rl}{RL}{Reinforcement Learning}
\newacronym{pomdp}{POMDP}{Partially Observable Markov Decision Process}
\anaacr{mdp}{MDP}{Markov Decision Process}
\anaacr{mpc}{MPC}{Model Predictive Control}
\newacronym{cem}{CEM}{Cross-Entropy Method}
\anaacr{mc}{MC}{Monte Carlo}
\anaacr{mi}{MI}{Mutual Information}
\newacronym{li}{LI}{Lautum Information}
\anaacr{nmc}{NMC}{Nested Monte Carlo}
\anaacr{sgd}{SGD}{Stochastic Gradient Descent}
\anaacr{svgd}{SVGD}{Stein Variational Gradient Descent}
\newacronym{rbf}{RBF}{Radial Basis Function}
\newacronym{boed}{BOED}{Bayesian Optimal Experimental Design}
\newacronym{elbo}{ELBO}{Evidence Lower Bound}
\anaacr{mse}{MSE}{Mean Squared Error}
\newacronym{sac}{SAC}{Soft Actor Critic}
\anaacr{mcts}{MCTS}{Monte Carlo Tree Search}
\anaacr{mtl}{MTL}{multi-task learning}

\newacronym{em}{EM}{Expectation Maximization}
\glsunset{em}
\newacronym{iid}{i.i.d.}{independent and identically distributed}
\glsunset{iid}
\newacronym{kl}{KL}{Kullback-Leibler}
\glsunset{kl}

\makeglossary

\DeclareUnicodeCharacter{1EF3}{\`y}

\newcommand{\balltaskname}{Tilted Pushing}
\newcommand{\ballhtaskname}{Tilted Pushing Maze}

\title{Active Inference for Robotic Manipulation}

\author{
    Tim Schneider \\
    Intelligent Autonomous Systems\\
    Technical University of Darmstadt\\
    64289 Darmstadt, Germany \\
    \texttt{tim@robot-learning.de} \\
    \And
    Boris Belousov \\
    Intelligent Autonomous Systems\\
    Technical University of Darmstadt\\
    64289 Darmstadt, Germany \\
    \texttt{boris@robot-learning.de} \\
    \And
    Hany Abdulsamad \\  
    Intelligent Autonomous Systems\\
    Technical University of Darmstadt\\
    64289 Darmstadt, Germany \\
    \texttt{hany@robot-learning.de} \\
    \AND
    Jan Peters \\
    Intelligent Autonomous Systems\\
    Technical University of Darmstadt\\
    64289 Darmstadt, Germany \\
    \texttt{mail@jan-peters.net} \\
}

\bibliography{bibliography/active_inference}
\bibliography{bibliography/general}
\bibliography{bibliography/mi}
\bibliography{bibliography/model_based_rl}
\bibliography{bibliography/active_inference_related_work}
\bibliography{bibliography/boed_related_work}

\glsdisablehyper

\begin{document}
    \maketitle

    \begin{abstract}
        Robotic manipulation stands as a largely unsolved problem despite significant advances in robotics and machine learning in the last decades.
        One of the central challenges of manipulation is partial observability, as the agent usually does not know all physical properties of the environment and the objects it is manipulating in advance.
        A recently emerging theory that deals with partial observability in an explicit manner is \glsxtrlong{ai}.
        It does so by driving the agent to act in a way that is not only goal-directed but also informative about the environment.
        In this work, we apply \glsxtrlong{ai} to a hard-to-explore simulated robotic manipulation tasks, in which the agent has to balance a ball into a target zone.
        Since the reward of this task is sparse, in order to explore this environment, the agent has to learn to balance the ball without any extrinsic feedback, purely driven by its own curiosity.
        We show that the information-seeking behavior induced by \glsxtrlong{ai} allows the agent to explore these challenging, sparse environments systematically.
        Finally, we conclude that using an information-seeking objective is beneficial in sparse environments and allows the agent to solve tasks in which methods that do not exhibit directed exploration fail.
    \end{abstract}

    \keywords{
        Model-Based Reinforcement Learning, Robotic Manipulation, Active Inference
    }

    \startmain

    \section{Introduction and Related Work}

    \begin{wrapfigure}{R}{0.25\textwidth}
        \centering
        \includegraphics[width=\linewidth]{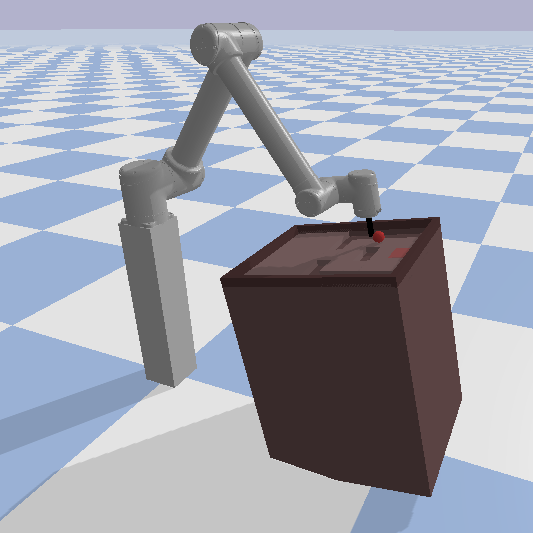}
        \caption{Robot using \glsxtrlong{ai} to solve a challenging manipulation task.}
        \label{fig:title_image}
    \end{wrapfigure}

    A common belief in cognitive science is that the evolution of dexterous manipulation capabilities was one of the major driving factors in the development of the human mind~\cite{macdougall1905significance}.
    Performing manipulation is cognitively highly demanding, forcing the actor to reason not only about the impact of its actions on itself but also about the impact on its environment.
    This inherent complexity leaves autonomous robotic manipulation a largely unsolved topic, despite significant advances in robotics and machine learning in the last decades.

    One of the central challenges of manipulation is partial observability.
    While we are manipulating an object, we rarely know all of its physical properties in advance.
    Instead, we must resort to inferring those properties based on observations and touch.
    To deal with this issue as effectively as possible, humans have developed various active haptic exploration strategies that they constantly apply during manipulation tasks~\cite{lacreuse1997manual}.

    A recently emerging theory from cognitive science that tries to explain this notion of constant active exploration is \acrfull{ai}~\cite{friston2010action}.
    \gls{ai} formulates both action and perception as the minimization of a single free-energy functional, called the \acrfull{vfe}.
    In doing so, \textcite{friston2015active} derive an objective function that consists of an extrinsic, goal-directed term and an intrinsic, information-seeking term.
    The combination of these two terms drives the agent to act in a way that is both goal-directed and informative, in that the agent learns about its environment through its actions.

    In this work, we show how \gls{ai} can be used to learn challenging robotic manipulation tasks without prior knowledge.
    For now, we assume that the environment is fully observable and only consider epistemic uncertainty\footnote{
        Epistemic uncertainty is the uncertainty the agent has over its model of the world.
        In contrast, aleatoric uncertainty is uncertainty over the agent's state.
    }.
    To implement \gls{ai} in practice, we use a neural network ensemble and deploy \glsxtrlong{mpc} for action selection.
    We show that agents driven by \gls{ai} explore their environments in a directed and systematic way.
    These exploratory capabilities allow the agents to solve complex sparse manipulation tasks, on which agents that are not explicitly information-seeking fail.

    Related to our approach is PETS~\cite{chua2018deep}, which also trains ensemble models for the transition and reward distributions and selects actions with a \glsxtrlong{cem} planner.
    The key difference to our approach is that PETS does not use an intrinsic term and instead greedily select the actions they predict to yield the highest reward.

    An approach similar to ours is \textcite{tschantz2020reinforcement}, who also tackle \gls{rl} tasks with \gls{ai}.
    The difference to our approach is that they use a different free energy functional used for planning and chose a different approximation of their intrinsic term, which requires them to make a mean-field assumption over consecutive states.
    They evaluate their approach on multiple \gls{rl} benchmarks, including \textit{Mountain Car} and \textit{Cup Catch}.

    \section{\texorpdfstring{\glsxtrlong{ai}}{Active Inference}}
    According to the \acrfull{fep}~\cite{friston2010action}, any organism must restrict the states it is visiting to a manageable amount.
    Mathematically, \gls{ai} implements this restriction as follows:
    Every agent maintains a generative model $p$ of the world and avoids sensations $o$ that are surprising, hence have a low marginal log-probability $\ln \p{o}$.
    Thus, the objective can be written as
    \begin{align}
        \label{eq:ai}
        \min_{\pi}\; -\ln \p{o}
    \end{align}
    where $o$ is generated by some external process that can be influenced by changing the policy $\pi$.

    The agent's generative model is assumed to consist of not only observations $o$, but also contain hidden states $x$, giving $\p{o} = \int \p{o, x} dx = \int \pc{o}{x} \p{x} dx$.
    To make \cref{eq:ai} tractable, we apply variational inference and obtain the \gls{elbo} using Jensen's inequality:
    \begin{align}
        -\ln \p{o}
        =
        -\ln \int \p{o, x} dx
        &=
        -\ln \int \frac{\qp{x}{\phi}}{\qp{x}{\phi}} \p{o, x} dx
        \leq
        \kl{\qp{x}{\phi}}{\pc{x}{o}} - \ln \p{o}
        \eqqcolon
        \vfesym \left( o, \phi \right)
    \end{align}
    where $\qp{x}{\phi}$ is the variational posterior, parameterized by $\phi$, and $\vfesym \left( o, \phi \right)$ is termed the \acrfull{vfe} in the \gls{ai} literature.

    Minimizing $\vfesym \left( o, \phi \right)$ w.r.t.\ the variational parameters $\phi$ corresponds to minimizing the \glsxtrshort{kl} divergence between the variational posterior $\qp{x}{\phi}$ and the true posterior $\pc{x}{o}$.
    In other words, by minimizing the \gls{vfe} w.r.t.\ $\phi$, the agent is solving the perception problem of mapping its observations to their latent causes.

    To facilitate planning into the future, the \gls{vfe} can be modified to incorporate an expectation over future states, yielding the \acrfull{efe}~\textcite{friston2015active}:
    \begin{align}
        \efesymP_{\pi} \left( \phi \right)
        &=
        - \Ex{\uqcp{o_{t+1:T}, x_{t+1:T}}{\pi}{\phi}}{
            \ln \p{o_{t+1:T}, x_{t+1:T}}
            - \ln \qcp{x_{t+1:T}}{\pi}{\phi}
        }
        \\
        &\approx
        -
        \underbrace{
            \Ex{\uqcp{x}{\pi}{\phi}}{
                \kl{\qcp{o}{x, \pi}{\phi}}{\qcp{o}{\pi}{\phi}}
            }
        }_{\text{intrinsic term (expected information gain)}}
        \underbrace{
            - \Ex{\uqcp{o}{\pi}{\phi}}{\ln \p{o}}
        }_{\text{extrinsic term}}
        \label{eq:efe}
    \end{align}
    where we omitted subscripts for readability and defined $\q[\phi]{o}[x] \coloneqq \p{o}[x]$, such that $\qsym_\phi$ and $\psym$ follow the same observation model.

    The minimization of the \gls{efe} w.r.t.\ the policy $\pi$ causes the agent to act in a way that maximizes both information gain and the extrinsic term.
    Here, the extrinsic term acts as an external signal that allows us to make the agent prefer or disprefer certain observations.
    While it is common in \gls{rl} literature to use a reward function to give the agent a notion of ``good'' and ``bad'' behavior, in the \gls{ai} framework, we define a prior distribution over target observations $\p{o}$ that we would like the agent to make.
    Note that by making the reward part of the observation and setting the maximum reward as target observation~\cite{tschantz2020reinforcement}, we can transform any reward-based task to fit into the \gls{ai} framework.

    \section{Method}

    In this work, we propose a model-based \glsxtrlong{rl} algorithm that uses \gls{ai} to efficiently explore challenging state spaces.
    Therefore, we assume that the environment is fully observable, governed by unknown dynamics $\PP{x_\tau}[x_{\tau - 1}, a_\tau]$ and provides the agent with a reward $\PP{r_\tau}[x_{\tau}, a_\tau]$ in every time step.
    We model both the dynamics and the reward with neural network conditioned Gaussians $\p{r_\tau}[x_{\tau}, a_\tau, \theta] \coloneqq \mathcal{N} \left( x_\tau \mid \mu^x_\theta \left( x_{\tau - 1}, a_\tau \right), \sigma^x I \right)$ and $\p{r_\tau}[x_{\tau}, a_\tau, \theta] \coloneqq \mathcal{N} \left( r_\tau \mid \mu^r_\theta \left( x_\tau, a_\tau \right), \sigma^r I \right)$, resulting in the following generative model:
    \begin{align}
        \label{eq:model_mdp}
        \p{x_{0:T}, a_{1:T}, r_{1:T}, \theta}
        =
        \p{x_0}
        \p{a_{1:T}}
        \p{\theta}
        \prod_{\tau = 1}^T
        \pc{r_\tau}{x_{\tau}, a_\tau, \theta}
        \pc{x_\tau}{x_{\tau - 1}, a_\tau, \theta}
    \end{align}

    Since the environment is fully observed, the only hidden variables are the neural network parameters $\theta$.
    Thus, we are left with the following minimization problem for selecting a policy $\pi \coloneqq a_{t+1:T}$ at time $t$:
    \begin{align}
        \min_{\pi}
        \efesymP_{\pi} \left( \phi \right)
        &\coloneqq
        -
        \underbrace{
            \Ex{\uqcp{\theta}{\pi}{\phi}}{
                \kl{\p{x_{t+1:T}, r_{t+1:T}}[\theta, \pi]}{\qcp{x_{t+1:T}, r_{t+1:T}}{\pi}{\phi}}
            }
        }_{\text{expected parameter information gain}}
        -
        \underbrace{
            \Ex{\uqcp{r_{t+1:T}}{\pi}{\phi}}{\sum_{\tau = t + 1}^T r_\tau}
        }_{\text{expected cumulative reward}}
        \label{eq:efe_real}
    \end{align}
    where we defined the observation preference distribution such that $\p{o_{\tau}} \propto e^{r_{\tau}}$, and defined $\q[\phi]{x, r}[\theta, \pi] \coloneqq \p{x, r}[\theta, \pi]$.
    Hence, by this definition, $\qsym$ and $\psym$ differ only in the marginal probability of the model parameters $\theta$.

    Similar to other methods utilizing \glsxtrlong{mpc}~\cite{chua2018deep}, by minimizing this objective function we select a policy that maximizes the expected cumulative reward over a fixed horizon.
    However, additionally we are maximizing the expected parameter information gain, driving the agent to seek out states that are informative about its model parameters $\theta$.
    This term causes the agent to be curious about its environment and explore it systematically, even in the total absence of extrinsic reward.
    The optimization of this objective can now theoretically be done by any planner that is capable of handling continuous action spaces.
    In this work, similar to \textcite{chua2018deep}, we use a variant of the \glsxtrlong{cem} to find an open loop sequence of actions $a_{t+1:T}$ that maximizes \cref{eq:efe_real}.

    A major challenge in computing $\efesymP_{\pi} \left( \phi \right)$ is that neither the intrinsic, nor the extrinsic term can be computed in closed form.
    While the extrinsic term can straightforwardly be approximated with sufficient accuracy via \glsxtrlong{mc}, the intrinsic term is known to be notoriously difficult to compute~\cite{mcallester2020formal}.
    Thus, instead of maximizing it directly, many methods maximize a variational lower bound of it~\cite{poole2019variational}.
    However, due to the high-dimensional nature of $\theta$, these approaches are too expensive to be executed during planning in real time.

    Hence, instead we propose to use a \glsxtrlong{nmc} estimator that reuses samples from the outer estimator in the inner estimator to approximate the intrinsic term:
    \begin{align}
        \label{eq:meth_nmc_est_reuse}
        \pr{\text{IG}}{(x, r), \theta}
        &\approx
        \underbrace{
            \frac{1}{n}
            \sum_{i=1}^{n}
            \ln \p{x_{i}, r_i}[\theta_{i}]
            -
            \ln
            \underbrace{
                \frac{1}{n}
                \sum_{\substack{k=1\\k\neq i}}^{n}
                \p{x_{i}, r_i}[\theta_{k}]
            }_{\text{inner estimator}}
        }_{\text{outer estimator}}
    \end{align}
    Although using the same samples $\theta_1, \dots, \theta_n$ in the inner estimator as in the outer estimator violates the \gls{iid} assumption, we found this reuse of samples to increase the sample efficiency substantially.
    Since this estimator only requires samples of $\theta$, we represent $\q[\phi]{\theta}$ by a set of particles $\theta_1, \dots, \theta_n$, making our model a neural network ensemble.

    \section{Experimental Results}

    \begin{wrapfigure}{R}{0.5\textwidth}
        \centering
        \begin{subfigure}[t]{0.45\linewidth}
            \centering
            \includegraphics[width=\linewidth]{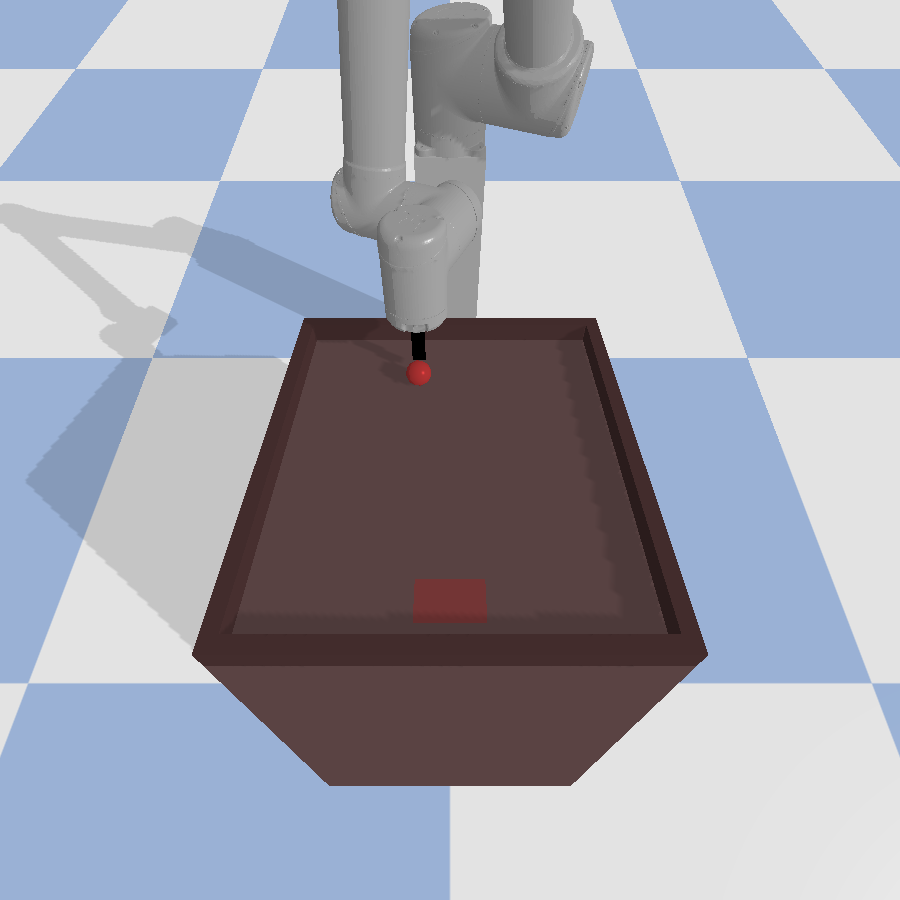}
        \end{subfigure}
        \hspace{0.5cm}
        \begin{subfigure}[t]{0.45\linewidth}
            \centering
            \includegraphics[width=\linewidth]{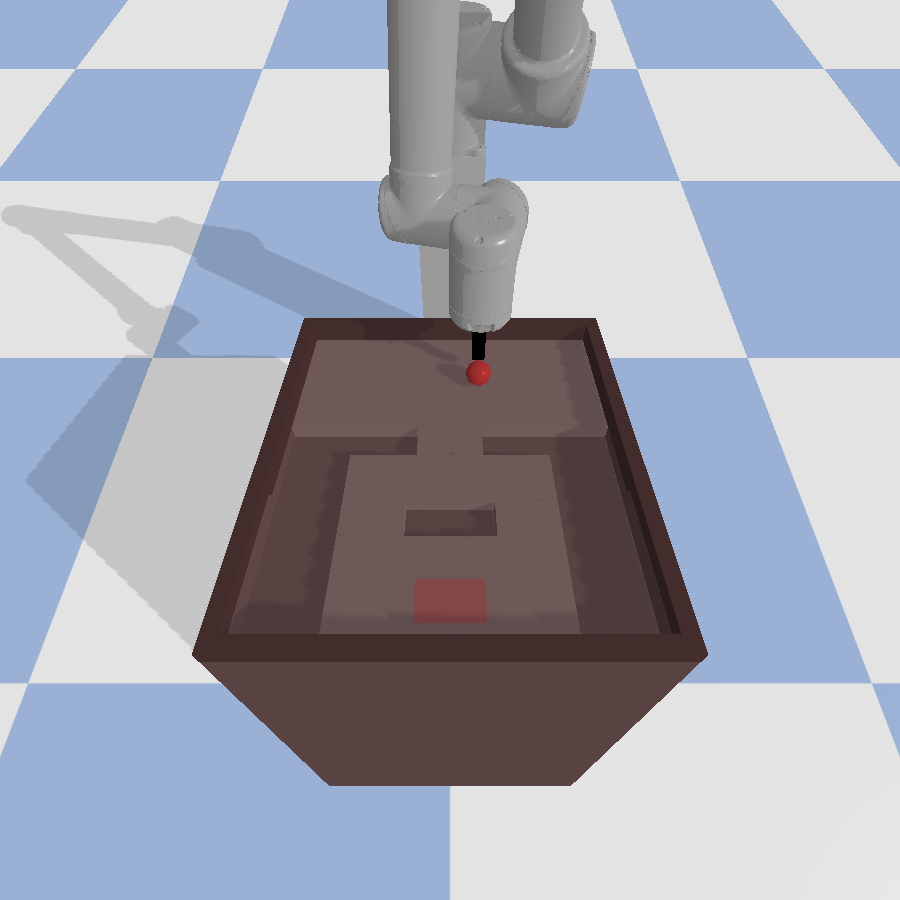}
        \end{subfigure}
        \caption{
            Visualization of the two environment configurations we test our methods on: \textit{\balltaskname} (left) and \textit{\ballhtaskname} (right).
            The target zone is marked in red.
        }
        \label{fig:tasks}
    \end{wrapfigure}

    A central feature that sets our method apart from other purely model-based approaches~\cite{hafner2019learning,chua2018deep} is the intrinsic term, that explicitly drives the agent to explore its environment in a systematic manner.
    To evaluate the exploratory capabilities of our method, we designed two hard-to-explore manipulation tasks: \textit{\balltaskname} and \textit{\ballhtaskname}.
    In both tasks, the agent has to push a ball up a tilted table into a target zone to receive reward.
    The agent can move the gripper in a plane parallel to the table and rotate the black end-effector around the Z-axis (Z-axis being orthogonal to the brown table and pointing up).
    As input, the agent receives the 2D positions and velocities of both the gripper and the ball, and the angular position and velocity of the end-effector.
    To add an additional challenge, in the \textit{\ballhtaskname} task we add holes to the table, that irrecoverably trap the ball if it falls in.
    For a visualization of these tasks, refer to \cref{fig:tasks}.

    There are two aspects make these tasks particularly challenging:
    First, the reward is sparse, meaning that the only way the agent can learn about the reward at the top of the table is by moving the ball there and exploring it.
    Second, balancing the ball on the finger and moving it around requires a fair amount of dexterity, especially given the low control frequency of 4 Hz\footnote{The computation of the intrinsic term is computationally heavy, limiting us to this rather low control frequency.} we operate our agent on.
    Once the agent drops the ball, it cannot be recovered, giving the agent no choice but to wait for the episode to terminate to continue exploring.
    Both of these aspects make solving these tasks with conventional, undirected exploration methods like Boltzmann exploration or adding Gaussian noise to the action extremely challenging.
    Consequently, the agent has to learn to balance the ball without receiving any extrinsic reward, purely driven by its own curiosity.

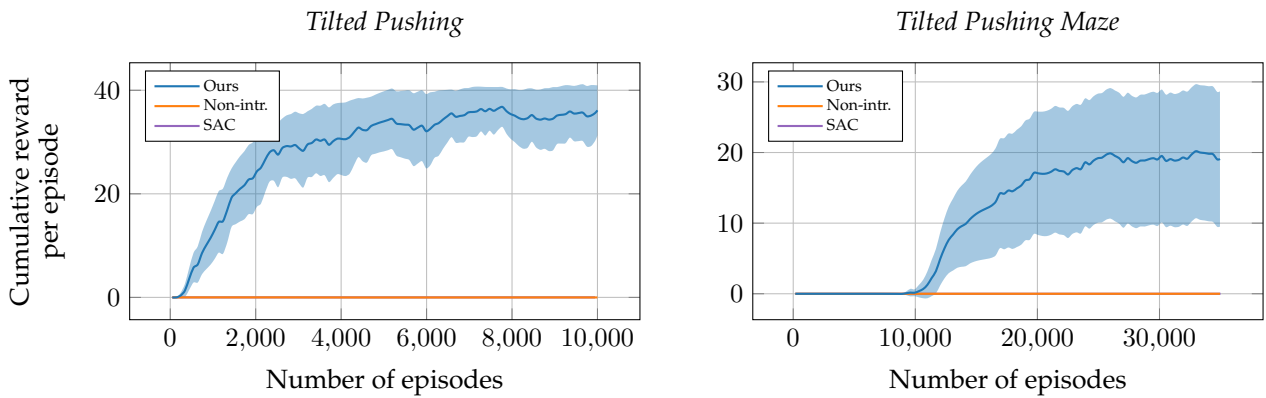
\begin{figure}[H]
        \centering
        \begin{tikzpicture}
            \begin{axis}
            [
                name=left,
                title={\textit{\balltaskname}},
                ylabel = {Cumulative reward\\per episode},
                rewplot, xtick={0,2000,...,10000}]
                \plotstdcom{data/ball_f05g20cla/sac_reward_test.csv}{pltPurple}{1}{SAC}{step}{mean_smooth}[std_smooth]
                \plotstdcom{data/ball_f05g20cla/ai/no_intrinsic_gan_rmsp_reward_test.csv}{pltOrange}{1}{Non-intr.}{step}{mean_smooth}[std_smooth][0.5]
                \plotstdcom{data/ball_f05g20cla/ai/mi_rmsp_reward_test.csv}{pltBlue}{1}{Ours}{step}{mean_smooth}[std_smooth][0.5]
            \end{axis}

            \coordinate (coord) at ([xshift=1.5cm]left.east);

            \begin{axis}
                [
                at={(coord)}, anchor=west,
                title={\textit{\ballhtaskname}},
                rewplot
                ]
                \plotstdcom{data/ball_f05g20h01cla/sac_reward_test.csv}{pltPurple}{1}{SAC}{step}{mean_smooth}[std_smooth]
                \plotstdcom{data/ball_f05g20h01cla/ai/no_intrinsic_gan_rmsp_reward_test.csv}{pltOrange}{1}{Non-intr.}{step}{mean_smooth}[std_smooth]
                \plotstdcom{data/ball_f05g20h01cla/ai/mi_reward_test.csv}{pltBlue}{1}{Ours}{step}{mean_smooth}[std_smooth][0.5]
            \end{axis}
        \end{tikzpicture}
        \caption{
            Cumulative per-episode reward for two different versions of our agent (one with intrinsic term, one without) and SAC on both variants of our environment.
            This graph displays the evaluation reward, which is obtained by rolling out the learned model without considering the intrinsic reward.
            Both non-intrinsic configurations and SAC failed to find the objective and converged to local minima.
        }
        \label{fig:exp_ball_res}
    \end{figure}

\begin{figure}[H]
        \centering
        \begin{tabular}{CCCCCC}
            Ours &
            \includegraphics[width=\linewidth]{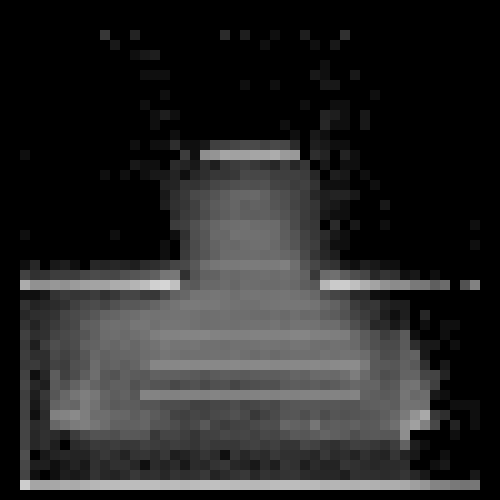} &
            \includegraphics[width=\linewidth]{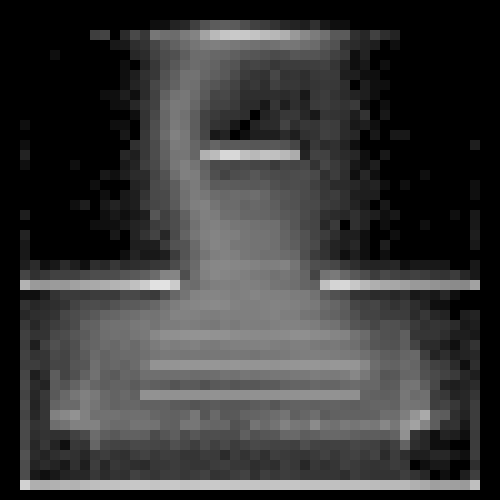} &
            \includegraphics[width=\linewidth]{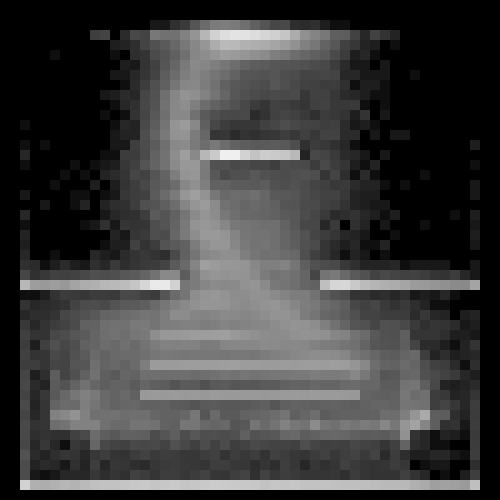} &
            \includegraphics[width=\linewidth]{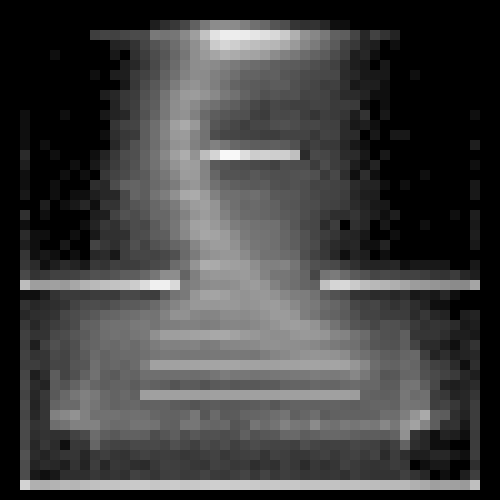} &
            \includegraphics[width=\linewidth]{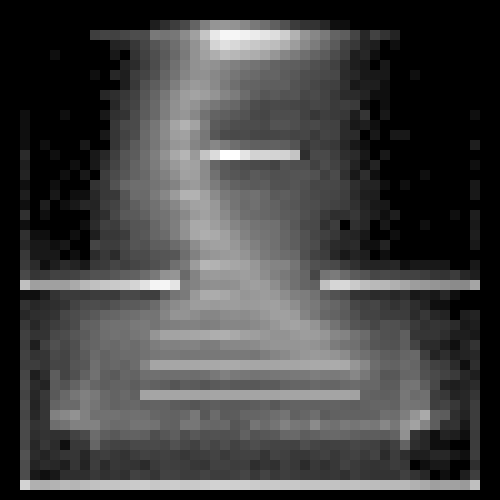}
            \\[1.2cm]
            Non-intrinsic &
            \includegraphics[width=\linewidth]{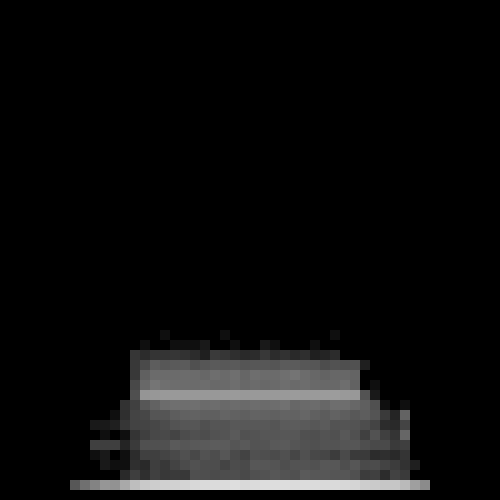} &
            \includegraphics[width=\linewidth]{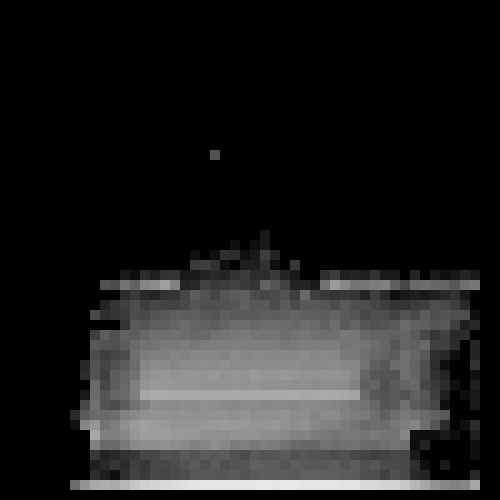} &
            \includegraphics[width=\linewidth]{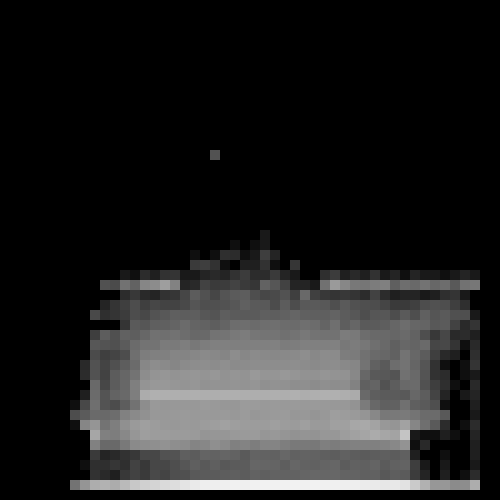} &
            \includegraphics[width=\linewidth]{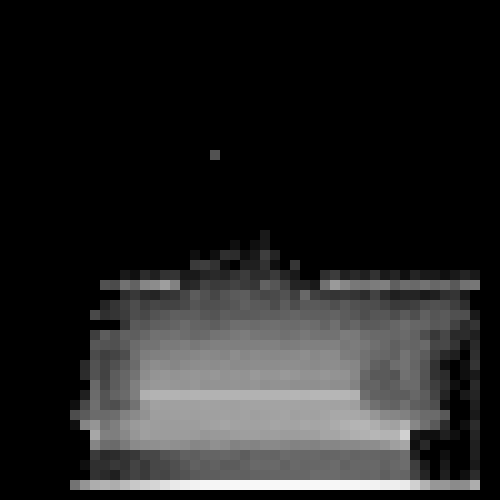} &
            \includegraphics[width=\linewidth]{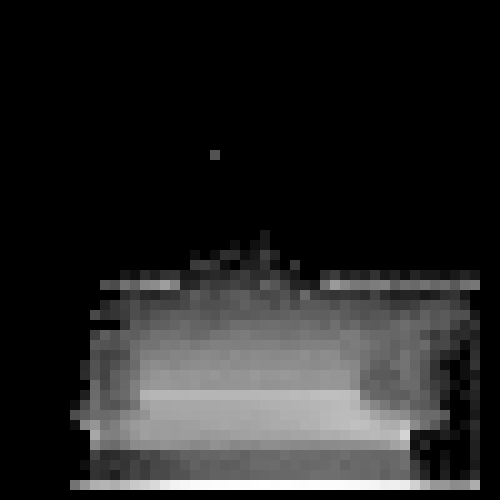}
            \\
            Episodes & 7000 & 14000 & 21000 & 28000 & 35000
        \end{tabular}\setlength\tabcolsep{6pt}
        \caption{
            Comparison of the states visited by our method and an agent using no intrinsic term, relying on Gaussian exploration instead.
            The brightness of each pixel indicates how often the ball has visited the respective point of the table at the given point in the training.
            The coordinate origin is at the bottom of each image, meaning that the images are rotated 180\textdegree compared to the top-down view in \cref{fig:tasks}.
            Each configuration was run once. %
        }
        \label{fig:exp_ballh_hist}
    \end{figure}

    As visible in \cref{fig:exp_ball_res}, our method is able to solve the \textit{\balltaskname}. 
    Both SAC~\cite{haarnoja2018soft} and our method without an intrinsic term fail to find the reward within 10,000 episodes.
    The holes of \textit{\ballhtaskname} make this environment significantly harder to explore, as the ball has to be maneuvered around two corners in order to reach the target zone.
    In this experiment, only our method finds the reward within 30,000 episodes.
    As can be seen in \cref{fig:exp_ballh_hist}, the reason for the bad performance of the non-intrinsic agent is its failure to explore the full state space.
    While our agent continues to systematically maneuver the ball around the holes in unseen locations, the non-intrinsic agent rarely passes the lower holes and leaves the upper half of the table unexplored.

    These experiment show that our method is able to systematically explore a complex, contact-rich environment with many dead-ends.
    Without any extrinsic feedback, our agents learned to balance the ball on the end-effector and systematically move it around the environment until the target zone was found.
    The sole reason for this behavior to occur in the first place is that our agents understood they could only explore the entire state space if they kept balancing the ball and move it to unseen locations.

    \section{Conclusion}

    In this work, we developed a method capable of applying \glsxtrlong{ai} to complex \glsxtrlong{rl} tasks.
    We evaluated our method in two challenging robotic manipulation task, both designed to be particularly hard-to-explore.
    Throughout our experiments, we showed that our method induces systematic exploration behavior and is capable of solving even the most challenging of these environments.
    Neither the non-intrinsic configurations nor the maximum entropy method SAC managed to solve the robotic manipulation tasks.
    Hence, we conclude that the information-seeking behavior of our agents is beneficial for solving challenging exploration problems with sparse rewards.

    Finally, in future work we plan to apply our method to a real robot and evaluate whether \glsxtrlong{ai} can be used in real robotic manipulation tasks.

    \printbibliography
\end{document}